# Non-Iterative Knowledge Fusion in Deep Convolutional Neural Networks

Mikhail Iu. Leontev[1,3], Viktoriia Islenteva[2], and Sergey V. Sukhov[*,2,3]

[1]S.P. Kapitsa Research Institute of Technology (Technological Research Institute) of Ulyanovsk State University, 4 Bld., 1 Universitetskaya Naberejnaya Str., 432000 Ulyanovsk, Russia

[2]Ulyanovsk State Technical University, 32 Severny Venets Str., 432027 Ulyanovsk, Russia

[3]Kotel'nikov Institute of Radio Engineering and Electronics of Russian Academy of Sciences (Ulyanovsk branch), 14 Spasskaya Str., 432011 Ulyanovsk, Russia

[*]Corresponding author: s_sukhov@hotmail.com

## Abstract

Incorporation of a new knowledge into neural networks with simultaneous preservation of the previous one is known to be a nontrivial problem. This problem becomes even more complex when new knowledge is contained not in new training examples, but inside the parameters (connection weights) of another neural network. Here we propose and test two methods allowing combining the knowledge contained in separate networks. One method is based on a simple operation of summation of weights of constituent neural networks. Another method assumes incorporation of a new knowledge by modification of weights nonessential for the preservation of already stored information. We show that with these methods the knowledge from one network can be transferred into another one non-iteratively without requiring training sessions. The fused network operates efficiently, performing classification far better than a chance level. The efficiency of the methods is quantified on several publicly available data sets in classification tasks both for shallow and deep neural networks.

## 1. Introduction and related work

Embedding new knowledge into a scheme of previously learned material is not a hard problem for biological neural networks. Artificial neural networks, despite their recent rapid development, still have a problem of incorporating new knowledge into an already learned structure. When new patterns are learned by a network, the new information may radically interfere with previously learned one, known as catastrophic interference (McCloskey & Cohen, 1989; Ratcliff, 1990). The goal of preserving previous knowledge can be achieved by joint training when the parameters of artificial neural network (ANN) are optimized by interleaving samples from new and old tasks (Caruana, 1998). The problem of incorporating new knowledge into pretrained ANN becomes more complicated if the data previously used for training this ANN are no longer available. Different approaches to tackle this problem were developed by various authors (Z. Li & Hoiem, 2016). Feature extraction (Donahue et al., 2014) and fine-tuning (Girshick, Donahue, Darrell, & Malik, 2014) are the common methods for training an existing network for a new task without requiring training data for the original tasks. More advanced techniques were also proposed recently (French, Ans, & Rousset, 2001; Z. Li & Hoiem, 2016).

A more complex situation for knowledge exchange occurs when different parts of knowledge are contained in different networks while the original training data are inaccessible or too large, or the data are sensitive/proprietary (for example, the information about the patients in hospitals). As a common approach, the integration of knowledge contained in different ANNs is performed through ensemble learning (Dietterich, 2000; H. Li, Wang, & Ding, 2017). In ensemble learning, the predictions produced by several ANNs are combined by weighted averaging or by voting to produce a single classification decision. A variety of ensemble learning methods have been introduced for obtaining a single prediction from several networks such as random forests (Breiman, 2001), Bayesian averaging (Domingos, 2000), stacking (Wolpert, 1992) etc.

The cost of using a combination of ANNs is that each of the constituent networks must be trained and then stored and the output of every network must be calculated during the prediction phase. Instead of constructing ensembles, there could be a necessity to save the storage and computational resources by combining several networks into a single one, transferring information from one network into another. Unfortunately, there is very limited literature devoted to the discussion of this particular problem. Most likely, this relates to the common interpretation of neural networks as black boxes that do not allow access to the internally stored information. Thus, so far, the exchange of knowledge between neural networks was considered close to impossible. Nevertheless, there have been several approaches that allow neural networks to train other networks. For this purpose, Zeng & Martinez (2000) used pseudo training set sampled from a distribution of the original training set. Other approaches, such as Model Compression or Model Distillation, were suggested in more recent research (Buciluă, Caruana, & Niculescu-Mizil, 2006; Hinton, Vinyals, & Dean, 2015; Papamakarios, 2015). In the Model Compression method, the authors use an ensemble of networks to label a large unlabeled dataset and then train single neural net on this much larger set. In Model Distillation, "soft targets" (softmax output under high temperature) were used for the training of a single network. However, all these methods require the presence of initial training data or large unlabeled dataset.

Recently, several approaches were put forward allowing to combine the knowledge contained in several shallow neural networks into a single one without the use of training data (so called, fusion of neural networks) (Akhlaghi & Sukhov, 2018; Smith & Gashler, 2017; Utans, 1996). In particular, it was shown that a simple approach of averaging weights of peer neural networks at periodic intervals is sufficient to facilitate the effective transfer of learned knowledge (Smith & Gashler, 2017; Utans, 1996). In these works, several networks with their weights located in the vicinity of one local minimum were averaged to produce single network that served as a better estimator to the input data. The weights located close to the different local maxima in this approach should be discarded (Utans, 1996). In our recent paper, we demonstrated that knowledge fusion can be performed even for networks with completely independent sets of weights (Akhlaghi & Sukhov, 2018). This method is based on a simple operation of summation of weights of two constituent networks to form one fused network. The method was tested on shallow networks in classification tasks and demonstrated decent performance (Akhlaghi & Sukhov, 2018). More details about this method will follow in Section 2.2.

Let us note that the above problem should not be confused with the term "knowledge transfer" (or "transfer learning"). In transfer learning, features of one pretrained network help in learning of a new problem for another network (Pan & Yang, 2010; Thrun & Pratt, 1998). Thus, transfer learning is more concerned with target task rather than maintaining knowledge about the source tasks. In this paper, we will use the transfer learning concept to help in generalizing the knowledge fusion methods for deep convolutional networks (see Sections 2.4).

The process of acquiring new knowledge is usually slow and iterative. While it is hard to speed up this process for biological neural networks, it can be done for ANNs, as one has access to all the parameters of neural networks. One of the currently known non-iterative methods used for the training of ANNs is Extreme Learning Machine (ELM) (Huang, Zhu, & Siew, 2004). In ELMs, part of the weights is assigned randomly, and the remaining weights are found noniteratively by a least-squares method. Of course, training data are still necessary for training of ELMs. In the following, we suggest new non-iterative methods for the fusion of knowledge contained in separate ANNs that do not require the original training data.

## 2 Theoretical background

### 2.1 Problem statement

We consider two feedforward neural networks $A$ and $B$ trained to classify a set of target classes $S_A$ and $S_B$. We assume that the networks are trained on separate sets of data and never see the examples from the classes belonging to another set. Each of the networks is characterized by its own set of parameters (weights) $\boldsymbol{\theta}_A$ and $\boldsymbol{\theta}_B$. Our intent is to combine these networks into a single one, which we call fused artificial neural network (fANN), such that it could classify the union of the all the target classes $S_A \cup S_B$ far better than chance level without having any training data or new training sessions (Figure 1). To be specific, we assume that both networks have the same architecture and we have all the information about their structure and weights but have no information about the original training data. Let us notice that the requirement of equal architecture is not obligatory and introduced only for convenience. In a general case of non-equal architectures, they can be made equal by introducing additional nodes and corresponding weights with zero values. Although we consider just two networks, the proposed approaches can be easily extended for several constituent networks.

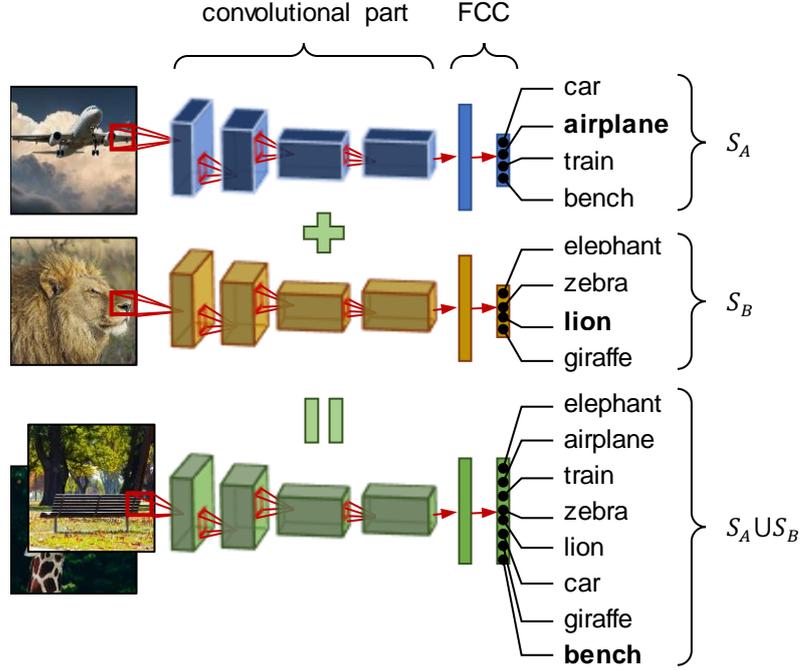

**Figure 1**. The schematics of the problem. Two deep convolutional networks each with its own knowledge are fused into a single one. The networks consist of a bottom convolutional part and top fully connected classifier (FCC). For nonintersecting set of classes $S_A$ and $S_B$, the output layer of the fused network is expanded to accommodate the union of classes $S_A \cup S_B$.

## 2.2 Weights Summation method

In our recent paper, we introduced the Weights Summation (WS) method (Akhlaghi & Sukhov, 2018) for the fusing of two or more neural networks. Here we briefly reiterate the ideas behind this method. In paper (Akhlaghi & Sukhov, 2018), we have shown that having an ensemble $M$ of ANNs trained to classify corresponding target classes, the information contained in those networks can be fused into one fANN through a simple summation of the corresponding weights

$$\theta_{F,i} = \sum_{m \in M} \theta_{m,i} . \qquad (1)$$

Surprisingly, in spite of the simplicity of this operation, the fANN obtained with the help of Eq.(1) can classify any entity from the joint set of classes with accuracy far better than chance level. Akhlaghi & Sukhov (2018) suggested the following explanation of WS method. The fused network can operate efficiently on all the target classes if activity of each neuron is not perturbed significantly. As typical activation functions (sigmoid, hyperbolic tangent etc.) usually restrict postsynaptic activity to be in a limited range, feeding the neuron with the collective presynaptic input leaning toward the right activity results in a strong correlation between the desired and the perturbed neuron activity. Of course, presynaptic connections deliver stochastic signal to each neuron. Consequently, the probability of having right activity can be quantified through a statistical approach. For an ensemble of two networks $M = \{A, B\}$ for a given feature vector $F = \{f_1, f_2, ..., f_d\}$ associated with class $c_A$ where $c_A \in S_A$ and $c_A \notin S_B$, each given feature $f_i$ together with the corresponding connection weights delivers presynaptic signals $\theta_A f_i$ and $\theta_F f_i$ in network $A$ and fANN, respectively. In this condition, the perturbed signal $\theta_F f_i$ drives a neuron with sigmoid activation function toward right activity, either excitation or inhibition, if $\text{sgn}(\theta_F f_i = \theta_A f_i + \theta_B f_i) = \text{sgn}(\theta_A f_i)$ where sgn denotes sign function. Having this equality, the probability of maintaining right activity in the neuron $P^{eq}$ will be

$$P^{eq} = p(\theta_F f_i > 0 | \theta_A f_i > 0) p(\theta_A f_i > 0) + p(\theta_F f_i < 0 | \theta_A f_i < 0) p(\theta_A f_i < 0). \qquad (2)$$

When $\theta_A f_i$ and $\theta_B f_i$ follow a normal distribution, one can estimate $P^{eq} \approx 0.75$ (Akhlaghi & Sukhov, 2018). Having $P^{eq} > 0.5$ guaranties right activity in each neuron if number of presynaptic connections tends to infinity. However, in practice, the number of presynaptic connections is limited. Having a large number of connections would be helpful to suppress fluctuations in neural activity from one sample to another, in a response to input feature vectors associated with the same target class. According to the law of large numbers, variation of unwanted fluctuations decreases by factor of $1/\sqrt{n}$ where $n$ is the number of independent presynaptic connections. This explains the fact that WS method generally performs better with the increase of the number of neurons in hidden layers (Akhlaghi & Sukhov, 2018).

The proposed explanation assumed that for WS method to work, one needs activation functions that restrict postsynaptic activity to be in a limited range (sigmoid, hyperbolic tangent etc.). However, below we will demonstrate that WS mechanism works well for unconstrained (e.g., ReLU) activation functions. That tells us that there are additional conditions providing the efficiency of WS method.

The presynaptic activity at arbitrary node in the fused network is determined by the following expression:

$$\theta_F f_i = \theta_A f_i + \theta_B f_i, \tag{3}$$

where $f_i$ is some feature belonging to $c_A$ class. In principle, the trained neural networks are robust under the perturbation of the weights (Blundell, Cornebise, Kavukcuoglu, & Wierstra, 2015). Thus, if the second term at the right hand side of Eq.(3) is significantly smaller that the first one, the fused network still will be able to perform classification independent on the activation function. To ensure the inequality $\theta_A f_i \gg \theta_B f_i$, several conditions should be met. First, the networks $A$ and $B$ should be well trained in a sense that they produce noticeable presynaptic activity in a presence of native feature and low activity in a presence of a foreign one. In this respect, the networks with random weights (e.g., Extreme Learning Machines (Huang et al., 2004; Schmidt, Kraaijveld, & Duin, 1992)) do not show good performance after fusion with WS method. In ELMs, the weights connecting input and hidden units are assigned randomly, so the majority of nodes would show similar activity for native and foreign input features $\theta_A f_i \approx \theta_B f_i$. We confirmed this statement to be valid for several sample classification problems, but do not provide those results in the present paper. Second, the mean of the weights probability distribution should be zero $\langle \theta_A \rangle = \langle \theta_B \rangle = 0$. This can be understood from the following. From a Bayesian perspective, the weights of a trained network $\theta_B$ are random variables taken from a posterior distribution $P(\boldsymbol{\theta}|\boldsymbol{x},\boldsymbol{y})$. For network $B$, feature vector $f_i$ represents another independent random variable. The product of independent random variables $\theta_B f_i$ can be rewritten as

$$\theta_B f_i \approx N_f \langle \theta_B \rangle \langle f_i \rangle. \tag{4}$$

Here $N_f$ is the number of elements in the feature vector and average $\langle \theta_B \rangle$ is taken over presynaptic weights of the layer. It follows from Eq.(4), that the term $\theta_B f_i$ will be close to zero if the expected value of weights distribution is zero $\langle \theta_B \rangle \approx 0$. The property $\langle \theta_B \rangle = 0$ is valid for common training techniques (Bellido & Fiesler, 1993; Blundell et al., 2015). In the meantime, one can imagine some exotic regularization techniques that make the average of weights probability distribution nonzero. Asymmetric probability distribution for weights appears also in training for certain classification tasks (Bellido & Fiesler, 1993). In those specific cases we expect the Weights Summation method fail. Also, WS would fail in neural networks, which try to mimic the networks of the brain, where the proportion of excitatory and inhibitory synapses is very unbalanced. For classification tasks used in the present research and for standard employed training techniques, the condition $\langle \theta_B \rangle = 0$ was always satisfied. Finally, WS method would fail if one overlaps neurons vital for the performance of both networks $A$ and $B$. Usually, ANNs are overparameterized with excess of weights, neuronal units, and layers. In this case the probability of overlap of essential nodes in Eq.(3) is low. With a decrease of number of neurons in a hidden layer $N_h$, the probability of overlapping of nodes essential for both networks increases. This is an additional factor of poor performance of WS method for small $N_h$ (see (Akhlaghi & Sukhov, 2018) and simulations in section 3.1). In principle, the performance of WS method can be improved if one would estimate the importance of each connection or neuronal unit. The method discussed in the next section attempts to perform such an estimation.

## 2.3 Elastic Weight Consolidation

Recently, several authors proposed a method for preventing catastrophic forgetting based on an approximation of the error surface of a trained neural network by a paraboloid in multi-dimensional space of weights (French & Chater, 2002; Kirkpatrick et al., 2017). The method selectively slows down learning on part of the weights important for the previously learned tasks. In other words, the modification of the weights of a retrained network occurs in a direction of minimum change of a loss function. Here we propose a modification of this method that can be used for the fusion of neural networks. Our method is based on the following idea. If we simultaneously change the weights of networks $A$ and $B$ so that they become equal with an additional condition of minimum change of loss functions $L_A(\boldsymbol{\theta})$ and $L_B(\boldsymbol{\theta})$, then the resulting network would have properties of both networks $A$ and $B$ (Figure 2). Following Kirkpatrick et al. (2017), we call this method Elastic Weight Consolidation (EWC).

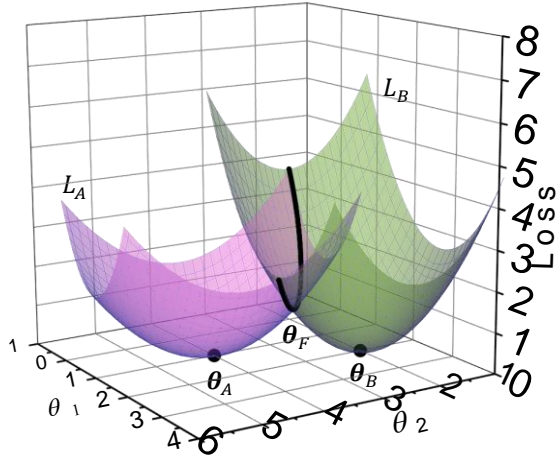

**Figure 2.** An illustration of the concept of Elastic Weights Consolidation. The surfaces of the loss functions $L_A$ and $L_B$ of neural networks $A$ and $B$ are approximated by paraboloids in the vicinity of optimal weights $\theta_A$ and $\theta_B$. The minimum value for the loss along the intersection of two paraboloids corresponds to the weight of the fused network $\theta_F$.

Following the original idea of French & Chater (2002), to control the change of the loss functions, we take the decomposition through a Taylor series with respect to the weights. Restricting ourselves to the second order terms, we have the following expression for network $A$:

$$L_A(\boldsymbol{\theta}) \approx L(\boldsymbol{\theta}_A) + \sum_i \left.\frac{\partial L}{\partial \theta_i}\right|_{\theta_i=\theta_{A,i}} (\theta_i - \theta_{A,i}) + \frac{1}{2}\sum_{i,j} \left.\frac{\partial^2 L}{\partial \theta_i \partial \theta_j}\right|_{\substack{\theta_i=\theta_{A,i}\\ \theta_j=\theta_{A,j}}} (\theta_i - \theta_{A,i})(\theta_j - \theta_{A,j}). \quad (5)$$

Here $\boldsymbol{\theta}_A$ are the weights of the trained network $A$, summation is performed over all the weights. Being a trained network, $L_A(\boldsymbol{\theta})$ achieves local minimum at $\boldsymbol{\theta} = \boldsymbol{\theta}_A$, thus, the first two terms in Eq.(5) can be disregarded. As the number of parameters in modern neural networks is in the millions, the calculation and storage of the whole Hessian matrix $\partial^2 L/(\partial \theta_i \partial \theta_j)$ becomes infeasible. Instead, following the approach of Kirkpatrick et al. (2017), we approximate the whole matrix by its diagonal terms:

$$L_A(\boldsymbol{\theta}) \approx \frac{1}{2}\sum_i F_{A,i} (\theta_i - \theta_{A,i})^2, \quad (6)$$

where $F_{A,i} \equiv \partial^2 L_A / \partial \theta_i^2$ are the diagonal components of the Hessian matrix also coinciding with Fisher information matrix (French & Chater, 2002; Pascanu & Bengio, 2013). The expression analogous to Eq.(6) can be written for network $B$ as well:

$$L_B(\boldsymbol{\theta}) \approx \frac{1}{2}\sum_i F_{B,i}\left(\theta_i - \theta_{B,i}\right)^2. \tag{7}$$

The coefficients $F_{A,i}$, $F_{B,i}$ represent the importance of weights $\theta_{A,i}$, $\theta_{B,i}$. Low values of $F_{A,i}$, $F_{B,i}$ mean that the corresponding weights can be safely changed without significantly affecting the loss.

The necessity of calculation of the second derivatives of a loss function appears in a number of techniques used in training of ANNs. Those techniques include the elimination of superfluous weights, estimation of confidence intervals, improving backpropagation algorithm etc. (Buntine & Weigend, 1994; Pascanu & Bengio, 2013). The exact calculation of Hessian matrix for the multilayer perceptron was performed by Bishop (1992). However, to use standard machine learning software packages, it is better to simplify the procedure of Hessian calculation. Luckily, in certain cases, the Hessian can be expressed only through the first derivatives. In the case of square loss function

$$L = \frac{1}{2}\sum_{p=1}^{N_p}\sum_{n=1}^{N_{out}}(y_n^p - t_n^p)^2, \tag{8}$$

the components of a Hessian matrix can be calculated as follows (French & Chater, 2002):

$$F_i = \sum_{p=1}^{N_p}\sum_{n=1}^{N_{out}}\left(\frac{\partial y_n^p}{\partial \theta_i}\bigg|_{\theta_i=\theta_{A,i}}\right)^2. \tag{9}$$

Here $y_n$ is the output from the $n$-th output unit of the network, and $t_n$ is the target value for the $n$-th output unit of the network. The summation in Eqs.(8),(9) is performed over all the output nodes $N_{out}$ and over all the training patterns $N_p$.

One another possibility of expressing Hessian through the first derivatives exists if the loss function corresponds to the negative logarithm of the likelihood of the training samples. This is the case of, for example, the cross-entropy error function. The second derivatives can be calculated as an expectation over the probability distribution of input patterns $x$ (Buntine & Weigend, 1994; Pascanu & Bengio, 2013)

$$F_i = \mathbb{E}_{p(x|\theta)}\left[\left(\frac{\partial L}{\partial \theta_i}\right)^2\right]. \tag{10}$$

In the following numerical experiments, we use both square and cross-entropy loss functions.

After the components of Fisher information matrix having been calculated, we need to find new set of weights $\theta_i = \theta_{F,i}$ that minimizes the combined loss function

$$L_F = L_A + L_B.$$

Correspondingly, the optimal weights $\boldsymbol{\theta}_F$ can be found from the equations

$$\frac{\partial L_F}{\partial \theta_i} = 0.$$

Taking into account Eqs.(6), (7), after straightforward calculations, we end up with the following expression for optimal weights $\boldsymbol{\theta}_F$ of the fused network:

$$\theta_{F,i} = \frac{F_{A,i}\theta_{A,i} + F_{B,i}\theta_{B,i}}{F_{A,i} + F_{B,i}}. \tag{11}$$

In the case when the weights $\boldsymbol{\theta}_A$, $\boldsymbol{\theta}_B$ are relatively close in the weights space, one can assume that $F_{A,i} \approx F_{B,i}$, and the optimal weights of the fused network $\boldsymbol{\theta}_F$ are just an average of corresponding weights of networks A and B: $\theta_F \approx (\theta_{A,i} + \theta_{B,i})/2$. This result was empirically found in (Smith & Gashler, 2017; Utans, 1996).

In principle, Eq.(11) solves the problem of knowledge fusion for two networks. However, in general case of networks A and B trained independently, the weights $\boldsymbol{\theta}_A$, $\boldsymbol{\theta}_B$ are not necessarily in the vicinity of each other. Thus, the approximation (5) would be not be precise enough. In attempt to bring the weights $\boldsymbol{\theta}_A$ and $\boldsymbol{\theta}_B$ closer to each other, one can try to "align" neural networks by rearranging nodes, taking into account that the nodes in any hidden layers of networks $A$ and $B$ can be arbitrarily permuted without the affecting the network's performance. Some simple methods of alignment of neural networks were considered previously (Ashmore & Gashler, 2015). Here, we find the optimal pairing of the hidden nodes of networks $A$ and $B$ by minimizing $L_F$. The problem in question can be stated as the following: for every node in a hidden layer of network $A$, one needs to find corresponding node in net $B$ in such a way that overall $L_F$ achieves minimum.

The contribution to the overall loss function $L_F$ from a pairing of a node $k$ of network $A$ and a node $l$ of network $B$ is

$$L_F^{kl} = \sum_i \left[ F_{A,i}^k \left( \theta_{F,i}^k - \theta_{A,i}^k \right)^2 + F_{B,i}^l \left( \theta_{F,i}^l - \theta_{B,i}^l \right)^2 \right] = \sum_i \frac{F_{A,i}^k F_{B,i}^l}{F_{A,i}^k + F_{B,i}^l} \left( \theta_{A,i}^k - \theta_{B,i}^l \right)^2 \qquad (12)$$

where the summation is performed over all the weights $i$ connected to the nodes $k$ and $l$. The total loss function $L_F$ can be found as a summation over all the pairs $\{k, l\}$:

$$L_F = \sum_{\{k,l\}} L_F^{kl}.$$

To find this optimal pairing, one needs to solve, so called, Assignment Problem, one of the fundamental combinatorial optimization problems, which has well established algorithms for its solution (Kuhn, 1955). The coefficients $L_F^{kl}$ compose the cost matrix that serves as input for the Assignment Problem algorithm.

In the case of multilayer feedforward network, the amount of all possible parings of the nodes in every layer dramatically increases. To accelerate the search for the (sub-) optimal solution, one can use the greedy algorithm in which two networks $A$ and $B$ are "zipped" together layer by layer starting from the deepest one. The Assignment Problem in this case is solved for every layer independently.

The overall algorithm of Elastic Weight Consolidation is outlined in Algorithm 1 panel.

---

**Algorithm 1.** Knowledge fusion using Elastic Weight Consolidation
1. Train two networks $\boldsymbol{A}$ and $\boldsymbol{B}$ on separate classes of data.
2. For every weight $\boldsymbol{\theta}_{A,i}$, $\boldsymbol{\theta}_{B,j}$ of networks $\boldsymbol{A}$ and $\boldsymbol{B}$, calculate diagonal components of Fisher information matrix $\boldsymbol{F}_{A,i}$ and $\boldsymbol{F}_{B,j}$ using, for example, Eqs.(9), (10).
3. **for** every hidden layer staring from the bottom one:
4.     For every unit in hidden layer calculate the components of the cost matrix $\boldsymbol{L}_{F,ij}$ according to Eq.(12).
5.     For the constructed cost matrix $\boldsymbol{L}_F$, solve the Assignment Problem.
6.     Perform the permutation of the nodes in the hidden layer of one of the networks according to the solution of the Assignment Problem.
7. **end for**
8. For every presynaptic weight of all the hidden units, calculate the weights of the fused network according to Eq. (11).
9. Concatenate the matrices of weights for the outer layers of networks $\boldsymbol{A}$ and $\boldsymbol{B}$.

---

According to Eqs.(9),(10), the components of Fisher information matrices $F_A$, $F_B$ used in EWC method should be calculated before throwing away training data. Thus, strictly speaking, EWC algorithm does not fully satisfy the requirements stated in the introduction that only the architecture and the weights of constituent networks are known. Although there are approaches of estimating the Hessian without the access to the original data (for example, by using noise as an input signal (French & Chater, 2002)), the quality of Hessian matrix obtained by these methods is questionable. Nevertheless, we can relax the original conditions and assume that components of Hessian matrix were computed at a training stage and were stored together with the weights for the following usage.

## 2.4 Fusion of deep convolutional neural networks

In our previous work, we demonstrated that the method of fusion of ANNs, namely, weights summation method, is suitable for the fusion of shallow neural networks (Akhlaghi & Sukhov, 2018). Moreover, WS method performs worse with the increase of the depth of the networks being fused (Akhlaghi & Sukhov, 2018). To extend fusion methods to deep convolutional networks and to simplify the calculations, we can employ the transfer of knowledge concept (Pan & Yang, 2010). According to transfer of knowledge concept, the weights of a network trained on one set of data can be used for the training on the data from a similar area. Applied to the problems in question, transfer of knowledge concept can be used as follows. Deep convolutional networks consist of a bottom convolutional part intended for the extraction of features from the raw input data, and a fully connected shallow network placed on top serving as a classifier (Figure 1). If the convolutional feature extractor is trained on a large enough set of classes, according to the transfer of knowledge concept, it can be used for the extracting of features for the data of new class of similar domain. At the beginning of training, two deep convolutional networks can be initialized to have the same bottom convolutional parts with shallow classifiers on top initialized by random weights. In the most trivial case, the bottom convolutional parts can be kept the same during training. In this case the fusion of deep convolutional networks is reduced to the fusion of shallow fully connected classifiers the same way as in paper by Akhlaghi & Sukhov (2018). In a more advanced case, deep ANNs can be fine-tuned by additional training of their convolutional parts. With the concept of knowledge transfer in place, fine-tuning would not result is significant weights modification. Thus, the weights of convolutional parts of networks $A$ and $B$ would be close in the weight space that is ideal for EWC method. For WS method, we should treat this situation little bit differently. For similar set of weights of the convolutional part, we cannot use the summation (3) as both terms in this expression would have similar values. Instead, the approach of (Smith & Gashler, 2017; Utans, 1996) would be ideal in this case and we can use weight averaging. However, the top fully connected parts still can be fused by weights summation.

## 3 Numerical experiments

### 3.1 Fusion of shallow neural networks

As a first example, we demonstrate the concepts of Weight Summation and Elastic Weight Consolidation on shallow neural networks using MNIST database of handwritten digits (LeCun, Bottou, Bengio, & Haffner, 1998). The whole dataset consists of 50000 training images of size 28 by 28 and 10000 test samples corresponding to 10 classes. We arbitrarily divided the whole dataset into two equal sets of classes containing 5 digits each. The intensity of every input image was normalized to be in the range [0,1]. Each network $A$ and $B$ was trained on data corresponding only to the target classes of particular network. Thus, before fusion, each network never saw the samples corresponding to the classes of other network and could not meaningfully classify corresponding images. The training algorithm was implemented in Keras 2.2 (Chollet, 2015) with the TensorFlow backend. Both networks $A$ and $B$ had an input layer with 784 nodes corresponding to the number of pixels in input images, one or more hidden layers with adjustable number of units, and output layer with 5 nodes corresponding to the number of classes. The units in hidden layers had ReLU activation functions. For output layer, two cases were considered: 1) sigmoid activation or 2) softmax activation. A square loss function was used together with a sigmoid activation in the output layer; with softmax activation, the cross-entropy loss was used. Before training, all the weights were initialized random from a normal distribution with zero mean with a standard deviation 0.05. Adam iterative method was used for the training (Kingma & Ba, 2014) with a learning rate 0.001 and other parameters set to the Keras' default values. In each experiment, a batch size of 200 was used. Early stopping was used while testing the accuracy on a validation set consisting of 12000 samples. We quantify the efficiency of classification of each network through accuracy (percent of correctly classified images). Simulations were repeated 10 times with random initialization of weights and with random subdivision of digits into classes to accumulate proper statistics.

      First, we notice that in the absence of hidden layers the fusion problem becomes trivial. In a case of separate classes, the fusion of networks $A$ and $B$ represents simple concatenation of matrices of weights of networks $A$ and $B$. It is necessary to prove that the fusion methods for multilayer networks

proposed in this paper work better than the linear classifier. Simulations show that the accuracy of classification of linear classifiers *A* and *B* is 0.96±0.01 for both sigmoid and softmax activations. After concatenation, the accuracy of classification of fused network with no hidden layers dropped to 0.774±0.009 for sigmoid activation and to 0.80±0.02 for softmax activation.

The classification accuracy of the fused network in a case when constituent networks *A* and *B* have one hidden layer is shown in Figure 3. The fusion performed by both WS and EWC methods is illustrated. One can see that for a number of neurons in a hidden layer exceeding 100, two-layer fused network outperforms the single layer one. From this we can conclude that linear classifier on its own cannot provide the type of accuracy observed when fusing two two-layer networks.

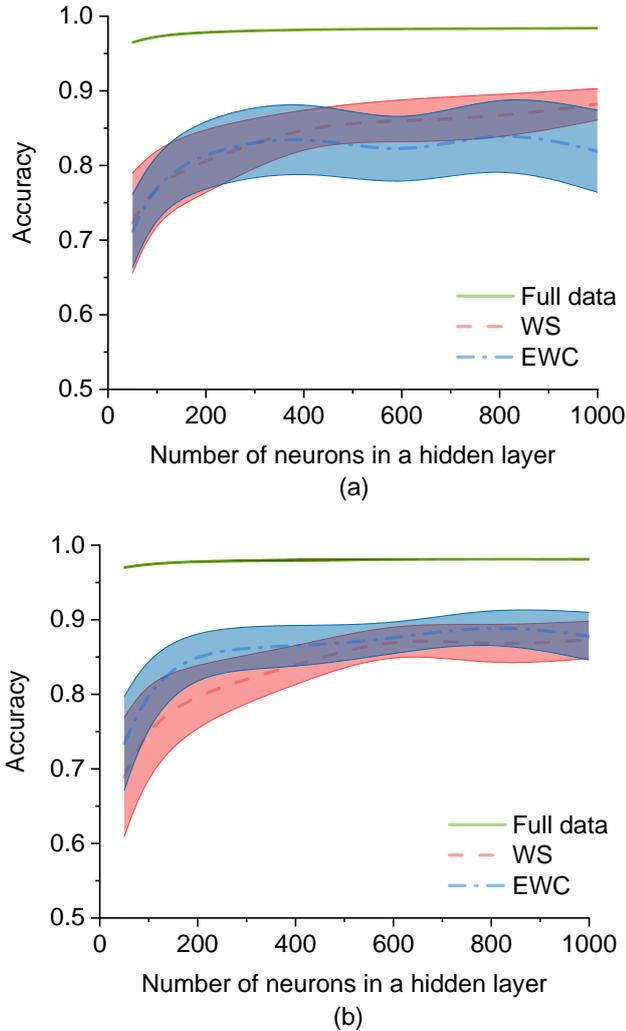

**Figure 3.** The accuracy of fANN as a function of number of neurons in a hidden layer. The cases of (a) sigmoidal output layer and square loss and (b) softmax output layer and cross-entropy loss are shown. Solid line (green) shows the accuracy of a network trained on the whole amount of data. Dashed line (red) shows the accuracy of fANN obtained by weights summation. Dot-dashed line (blue) shows the accuracy of fANN obtained by elastic weight consolidation. Shaded areas show standard deviation of the accuracy calculated over 10 experiments.

One can see from Figure 3, that having perturbations to the node activity in WS method results in higher fluctuations in the error rate of classification in fANN. In (Akhlaghi & Sukhov, 2018), we observed much more significant decrease of error rate in fANN by increasing the number of hidden nodes as compared to Figure 3. The difference with the present case is the absence of counter-examples during training. In our paper (Akhlaghi & Sukhov, 2018), both networks *A* and *B* saw the samples from all the classes during training. Although, the constituent networks could not classify the items coming from foreign classes, they could learn to distinguish them as "not-being-any-of-native" ones.

Surprisingly, EWC method shows very similar performance compared to WS both in error rate and in its variance (Figure 3). WS shows the power of statistical methods: in spite of heavier calculations required for EWC method, it cannot outperform WS method in wide range of parameters.

For WS method, the computation time grows linearly with number of neurons in a hidden layer $N_h$. Because of the nature of Hungarian algorithm used for the solution of the Assignment Problem in EWC method, the computation time for this method grows as $N_h^3$ and the method becomes very slow for $N_h \approx 1000$. However, the solution of Assignment Problem is vital for the performance of EWC method; without this additional optimization, EWC method resulted in very low accuracy. For example, for a neural network with 800 hidden units, the accuracy of fANN is at the level of 0.68±0.05.

The advantages of EWC method start to appear with the increase of the depth of a network. Figure 4 shows the accuracy of classification of MNIST data by fANNs obtained by WS and EWC methods. One can see that while the accuracy of WS fANN drops with the increase of a number of hidden layers (the result also found in (Akhlaghi & Sukhov, 2018)), the EWC fANN maintains its accuracy.

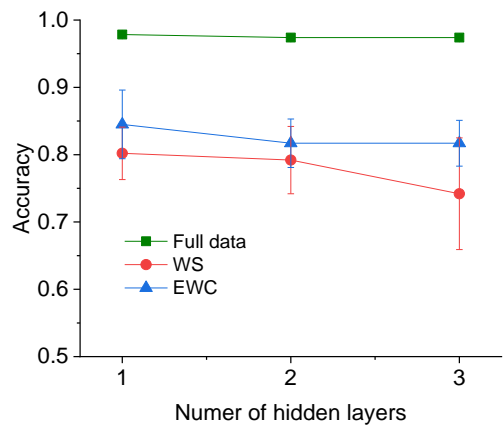

**Figure 4.** The accuracy of fANN as a function of number of hidden layers. fANNs is obtained by WS (circles, red curve) and EWC (triangles, blue curve) methods. Squares (green curve) shows the accuracy of a network trained on the whole amount of data. The fused networks had 200 neurons in every hidden layer with ReLU activation function. Output layer had softmax activation function. Loss function was categorical cross-entropy. Training was performed with early stopping, L2 regularization is used for every layer with coefficient 0.001. Adam optimizer was used for training. Batch size was 200.

### 3.2 Fusion of neural networks with common convolutional part

In this section, we demonstrate the generalization of the fusion methods for deep convolutional networks. As we saw in the previous section, current fusion methods decrease their performance with an increase of the depth of the neural networks. Thus, fusion of deep neural networks with completely independent sets of weights could be problematic in the discussed methods. To make fusion methods work, we employ the transfer of knowledge concept (Pan & Yang, 2010). This concept assumes that it is not necessary to train any deep network from a scratch. Instead one can use a part of other network (usually, its bottom convolutional part) trained on data of similar category. The convolutional part plays a role of extractor for the essential features that can be used by the classifier of top of the network to make a final prediction. In this section we consider a simpler example in which the convolutional parts of two constituent networks are the same.

For this example, we used CIFAR-10 dataset of 60000 color images 32 × 32 pixels each subdivided into 10 main categories (airplane, automobile, bird, cat, deer, dog, frog, horse, ship, truck) (Krizhevsky & Hinton, 2009). The classes are completely mutually exclusive. All the training was performed in Keras 2.2 with the Tensorflow backend. The images were normalized to have intensity of every channel in the range [0,1].

First, the six-layer convolutional network (see Table 1) was trained on a whole amount of data to be able to classify all 10 classes of objects of CIFAR-10 dataset. During the training, 10% of images were dedicated to the validation. Adam optimized was used with 0.0001 learning rate and $10^{-6}$ learning rate decay after every iteration. Training was performed with 128 images batch size over about 50 epochs with early stopping.

**Table 1**
The configurations of the convolutional network used for the training on CIFAR-10 data. The convolutional layer parameters are denoted as "conv(receptive field size)-(number of channels)". All hidden layers used the rectification (ReLU) non-linearity (Krizhevsky, Sutskever, & Hinton, 2012). Maxpooling is performed over a 2 × 2 pixel window, with stride 2. The probability of ignoring nodes in dropout is indicated in parentheses.

| input (32 × 32 RGB image) |
| --- |
| 6 weight layers |
| conv3-32 |
| conv3-64 |
| maxpool |
| dropout(0.25) |
| conv3-128 |
| maxpool |
| conv3-128 |
| maxpool |
| dropout(0.25) |
| flatten |
| fully connected 1024 |
| dropout(0.5) |
| fully connected 10 |
| softmax |

After the training on the whole amount of data, the outer two-layers of fully connected classifier were discarded; the weights of the bottom convolutional part were frozen and reused for further simulations. For networks $A$ and $B$, we supplemented the frozen convolutional part with new fully connected classifier containing one hidden layer with 1024 neurons and output layer with 5 neurons. In the transfer of knowledge ideology, the bottom part served as a universal feature extractor that provided extracted features to the fully connected classifier.

The whole CIFAR-10 dataset then was arbitrary divided into two parts containing 5 classes each. Networks $A$ and $B$ were trained on these separate datasets. Trained networks were fused into a single fANN using WS and EWC methods. In fANN, the bottom convolutional part remained the same as in networks $A$ and $B$. The procedures of WS and EWC were applied only to the top fully connected classifiers. The whole process was repeated 10 times with random initialization of weights and with random subdivision of classes to accumulate proper statistics. The results of numerical experiments are summarized in Figure 5. As in experiments with the fusion of shallow neural networks, WS and EWC methods show similar accuracy. Fused ANNs also show ≈ 10% worse performance as compared to the network trained on the whole amount of data.

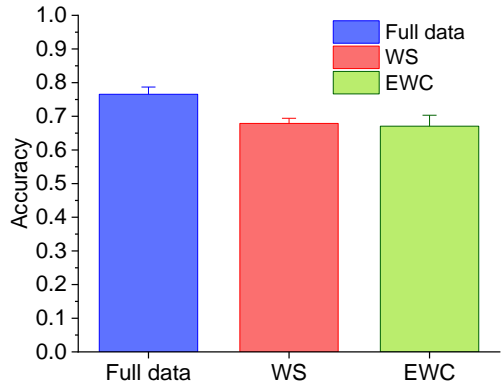

**Figure 5.** The accuracy of the classification of CIFAR-10 test data for a network trained on a full amount of data and for fANNs obtained by WS and EWC methods.

### 3.3 Fusion of deep neural networks

For the final example, we consider the fusion of deep convolutional networks. The networks in this example were trained on a set of natural images obtained from ImageNet project (Russakovsky et al., 2015) (www.image-net.org). Considering the variety of the objects indexed in the project, we can construct the networks trained to classify either similar or dissimilar classes of objects. In particular, for the current experiment, two "indigenous" neural networks were trained to classify 11 African and 11 Australian animals (see Table 2). The third, "urban", neural network was trained to classify 11 objects pertinent to the urban life (Table 2). Each image of the training dataset contained one or a group of animals (or urban objects), which could be located at different positions within the frame and could be imaged from different angles. The photographs could also contain the depiction of the habitat of the animals (trees, bushes, grass, lakes etc.) or different anthropogenic artefacts.

As in previous examples, it is assumed that the original trained networks observed only the objects of their own classes and never saw the objects from the classes of other networks. The aim of the exercise is to embed the knowledge contained in one neural networks into another one in such a way that the resulting network can recognize the objects (e.g., animals or household item) from the previously unknown category even without ever "seeing" the objects from those categories before.

**Table 2**

Classes of objects (synsets) of ImageNet database used for training three neural networks.

| African animals | | | Australian animals | | | Urban objects | | |
|---|---|---|---|---|---|---|---|---|
| name | Synset index | No. of images | name | Synset index | No. of images | name | Synset index | No. of images |
| gazelle | n02423022 | 1384 | dingo | n02115641 | 1262 | laptop | n03642806 | 1387 |
| African elephant | n02504458 | 2277 | koala | n01882714 | 2469 | airplane | n02691156 | 1434 |
| zebra | n02391049 | 1474 | echidna | n01872401 | 1336 | park bench | n03891251 | 1233 |
| chimpanzee | n02481823 | 1502 | wallaby | n01877812 | 1599 | desk | n03179701 | 1366 |
| gorilla | n02480855 | 1915 | platypus | n01873310 | 1078 | telephone | n04401088 | 1321 |
| hippopotamus | n02398521 | 1391 | wombat | n01883070 | 1222 | television set | n04405907 | 1268 |
| lion | n02129165 | 1795 | Tasmanian devil | n01884104 | 1347 | chair | n03001627 | 1460 |
| cheetah | n02130308 | 1427 | flying fox | n02140049 | 1202 | train | n04468005 | 1312 |
| ostrich | n01518878 | 1393 | cassowary | n01519563 | 1347 | automobile | n02958343 | 1307 |
| rhinoceros | n02391994 | 1496 | emu | n01519873 | 1212 | building | n02913152 | 1421 |
| giraffe | n02439033 | 1256 | giant kangaroo | n01877606 | 1150 | clock | n03046257 | 1615 |

In their 2014 paper, Simonyan and Zisserman demonstrated that the accuracy of classification in neural networks can be significantly improved if one increases the number of layers up to 16-19 (Simonyan & Zisserman, 2014). Thus, for the architecture of networks *A* and *B*, we chose VGG-16 model (Simonyan & Zisserman, 2014) included in Keras version 2.2 with its weights pre-trained on 1000 ImageNet classes. Those 1000 classes include large variety of objects, but, still, do not contain all the objects from Table 2. However, within the concept of knowledge transfer, the pretrained convolutional part of VGG-16 network is a good starting point for the training on previously unknown objects.

In the pretrained VGG-16 network, we kept only its bottom convolutional part that we used as universal feature extractor. On top of this convolutional part, we placed a custom classifier consisting of one hidden and one output layer with 11 neurons corresponding to the number of classes of the networks being fused. Dropout was used for the weights connecting hidden to output nodes of the classifier. ReLU activation function was used for the hidden layer, softmax activation was used for the output layer. For the hidden layer of the classifier, we tried different number of neurons starting from 128 and up to 1024. As the performance of the fused networks in our experiments generally increased with the increase of the number of neurons in the hidden layer, in the following we provide the results only for the network with 1024 neurons.

The input to VGG-16 network has to be a fixed-size 224 × 224 RGB image that was cropped out of a center of input raw images. Note, that not all the images provided by ImageNet had corresponding bounding boxes, so we did not use the information from the bounding boxes for the cropping of the images. Augmentation of dataset was not used. All the images were preprocessed by scaling the intensity of every image to the range [0,1]. 15% of the images were dedicated to validation and another 15% were used for the test. For the training, we used 64 picture batch, and the training was performed with Adam optimizer on a computer with GPU support (Ubuntu 16.04 x64, Intel core i3, 6GB RAM, Nvidia GeForce GTX 1050 Ti 4GB).

The training procedure for the constituent networks was as follows. First, the bottom convolutional part was frozen, and the top classifier was initialized with random weights and trained with small learning rate ($10^{-5}$) over 8 epochs. Then we froze the weights of the classifier and unfroze the top block (3 layers) of VGG-16 convolutional part. This block was trained over 8 epochs with the same small learning rate. Such a training procedure resulted in networks *A* and *B* different not only in their top fully connected parts, but also in their convolutional layers. As a result, the networks trained to classify African and Australian animals achieved 0.951±0.005 and 0.933±0.004 accuracy, correspondingly. The network trained to classify urban objects achieved 0.955±0.003 accuracy. Figure 6 shows corresponding confusion matrices for the trained networks. As one can see, the main source of errors for the African network (Figure 6a) is mixing between chimpanzees and gorillas. The Australian network, in its turn, sometimes mixes kangaroo for wallaby (Figure 6b). No noticeable mixing of classes is observable for the urban neural network.

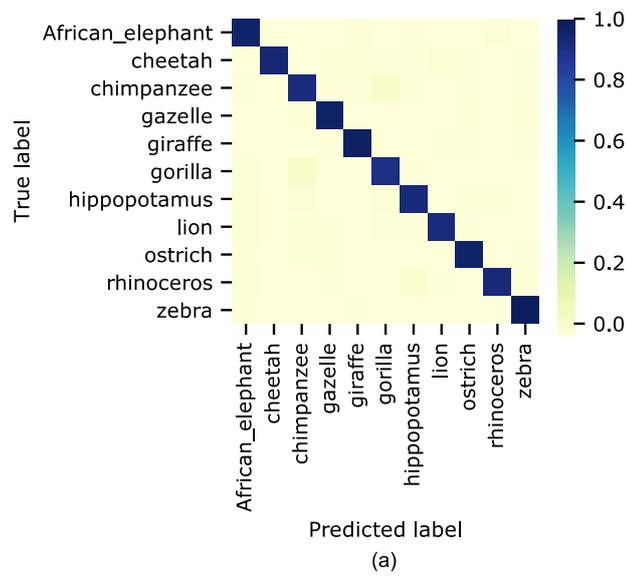

(a)

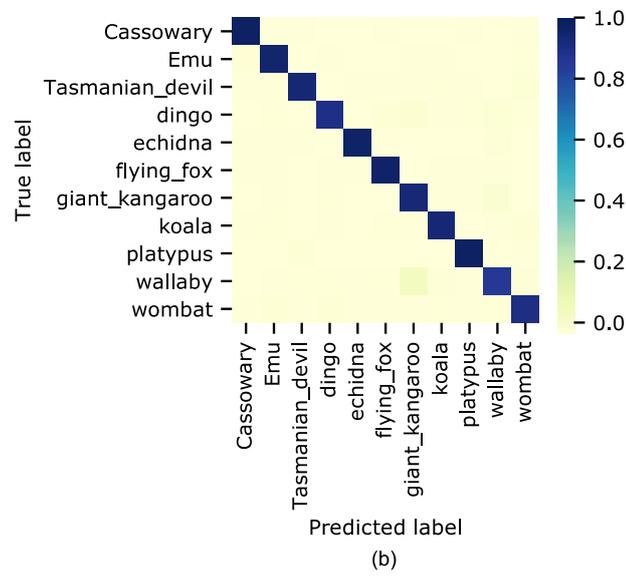

(b)

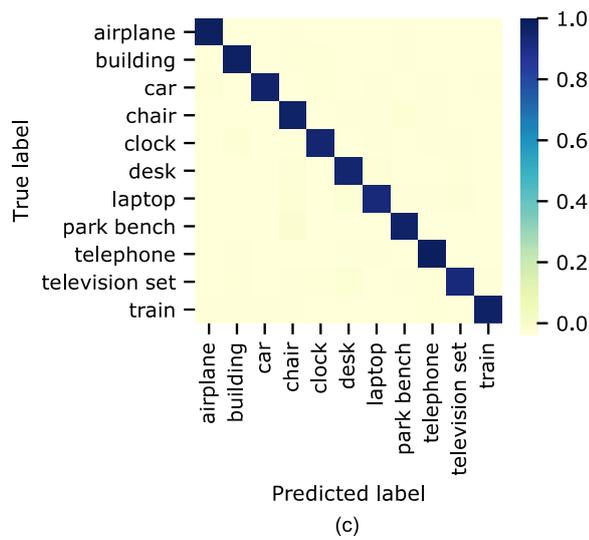

(c)

**Figure 6.** Confusion matrices for the constituent neural networks. The networks were trained to classify (a) African and (b) Australian animals and (c) urban objects.

The knowledge from the constituent networks was then fused by WS and EWC methods. Contrary to the previous examples, here we fuse not only shallow fully connected parts, but also bottom convolutional parts. As discussed in Section 2.4, for WS method we used slightly different procedure as before. As the weights of the convolutional parts of both networks originated from the same set of pretrained weights, during fusion by WS method, those weights were averaged according to approach adopted by Smith & Gashler (2017), Utans (1996). The weights of the classifiers on top of networks were obtained from an independent set of weights and thus were summed up as in all previous examples. The procedure of calculation of optimal weights for EWC method was the same independently on the location of the weights (Eq.(11)).

First, we fused two "indigenous" networks to check the accuracy of fusion on objects of similar categories. The accuracy of classification of fANNs obtained by WS method was 0.759±0.007, EWC resulted in 0.733±0.028 accuracy, which is a rather good result considering the fact that before fusion the networks never saw the objects from another category. The confusion matrix of fANN obtained by WS method for the joint set of classes is shown in Figure 7. Inherited from the original networks, chimpanzees are still confused with gorillas, kangaroos are confused with wallabys. Further analyzing the confusion matrix, one can notice that the classification of ostriches and emus is somewhat mixed. This happens because the original networks never learned the features that distinguish those very similar birds. Interestingly, lions are confused with dingos, platypuses are confused with hippopotami, as fANN finds that those animals have similar features. Similar confusion matrix for the fANN obtained by EWC method is shown in Supplementary Materials.

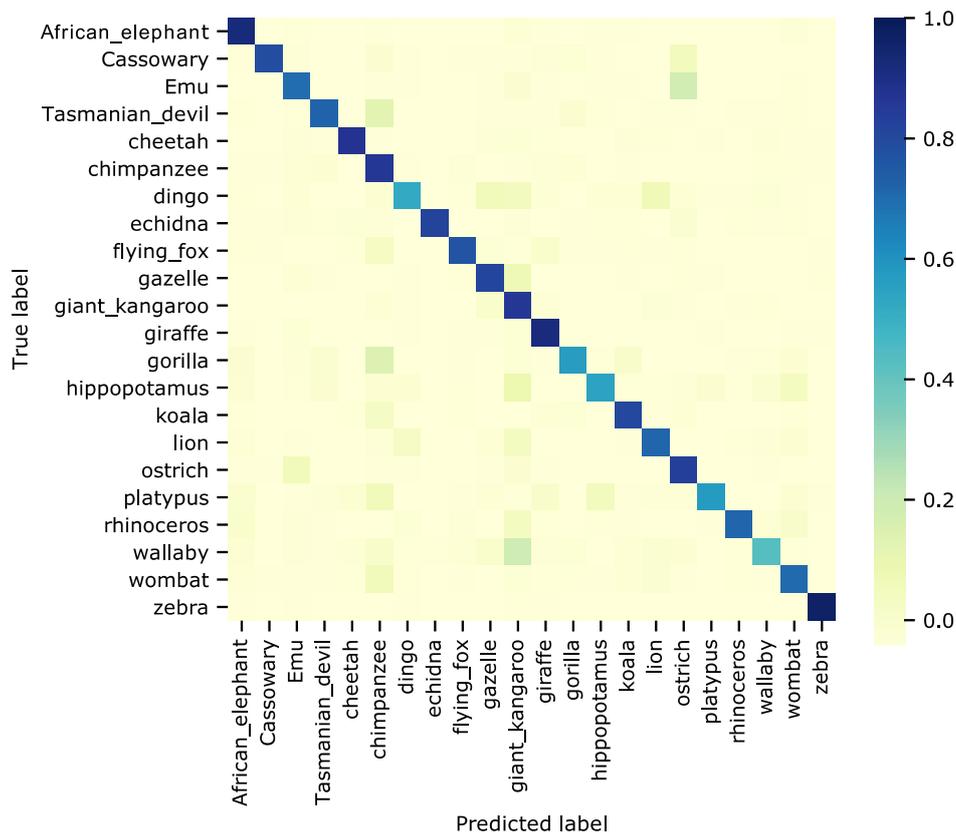

**Figure 7**. The confusion matrix for fANN obtained from 'African' and 'Australian' ANNs by WS method.

Second, we fused the networks with rather different prior 'experiences': network *A* was trained to classify African animals and network *B* was trained on urban objects. The accuracy of fANN achieved 0.808±0.016 for WS method and 0.835±0.008 for EWC method, which is noticeably higher than in previous case. The higher accuracy is achieved, because the objects from two sets of classes have very different features and their misinterpretation was less probable. Figure 8 shows the corresponding confusion matrix. WS method shows some spurious misinterpretations when certain African animals are taken for a one of the classes of urban network ('park bench' in this particular case). The confusion matrix for EWC method is shown in Supplementary Materials.

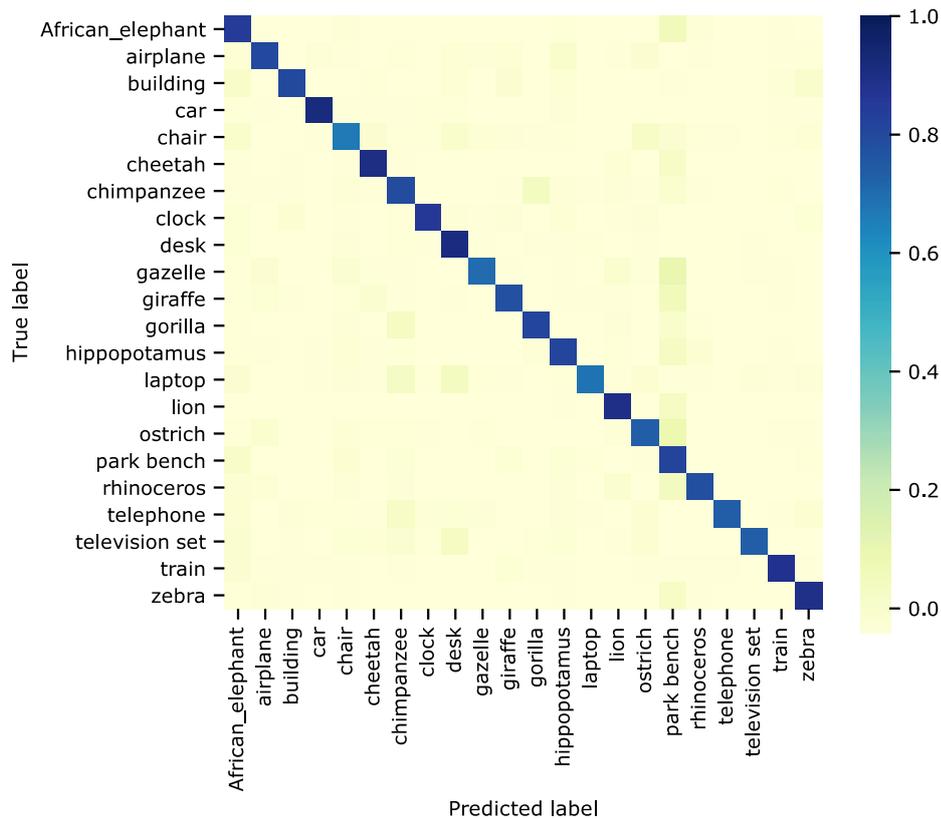

Figure 8. The confusion matrix for fANN obtained from 'African' and 'urban' ANNs by WS method.

## 4 Discussion

All the above numerical experiments show that fused neural networks demonstrate somewhat less accuracy than a network trained on the whole amount of data. Partially this can be attributed to the nature of fusion methods that incorporate some degree of stochasticity (WS) or use a number of approximations (EWC). However, even with future more advanced methods of fusion, one should not expect to achieve the accuracy of the network trained on a whole amount of data. The reason is that in the case of the whole dataset, the network sees many more counter-examples to correct its behavior. Without these counter-examples, the constituent networks could disregard important features that could distinguish the items of native class from similar items of the counterpart network (ostrich-emu misinterpretation for example). Thus, future fusion methods should take more active role in amplifying features that distinguish objects of two similar classes of constituent networks.

One another noticeable feature of fANN is high variability of the accuracy from one test to another. This variability is partially explained by different subdivision of original datasets into two parts from one test experiment to another as in experiments with MNIST and CIFAR-10 data. However, even with the same subdivision (as in experiment with ImageNet data), the variability of accuracy of fANN is 2…4 times larger than the one of constituent networks. Future fusion methods should aim to decrease the variability of efficiency of fANN.

The experiments performed in this paper show that Weights Summation method works beyond the limitations originally determined by Akhlaghi & Sukhov (2018). In particular, this method performs well with many activation functions (sigmoid, ReLU, softmax), with various regularization techniques ($L_2$, dropout), and with various loss functions (mean square, cross-entropy).

Surprisingly, a simple method of weights summation performs on a par or even better than more involved EWC method. It is determined by a number of approximations and simplifications used in the derivation of EWC. First, in EWC method, the real error surface of neural networks is approximated by a paraboloid as in Eq.(5). This approximation can suffice in application to the problem of catastrophic interference (Kirkpatrick et al., 2017) where the solution for a new problem is explicitly sought in the vicinity of the old one. In the case of pretrained networks, their weights are already predefined, and they can be far from each other in the weight space, which makes the paraboloid approximation insufficient. Second, in the EWC approach, the entire Hessian matrix is approximated by its diagonal part (Eqs.(6), (7)). One can expect that using nondiagonal terms of Hessian matrix can improve the fusion results. However, storing nondiagonal terms would require additional computer memory and additional computations. Third, all classes have their own error surface with its unique shape. In the presence of several classes, EWC approach averages out all these unique shapes and replaces them with a single effective parameter. This averaging harms the original idea of finding nodes (neurons) pertinent to a particular problem and restricting their change (Kirkpatrick et al., 2017). By averaging the unique error surfaces, we bring EWC method closer to a simple $L_2$ regularization technique. Finally, EWC implicitly assumes that modification of weights $\boldsymbol{\theta}$ of constituent networks does not change the posterior probability distributions $p(\boldsymbol{y}|\boldsymbol{x},\boldsymbol{\theta})$. This assumption may hold for small changes of weights, but generally is not correct and introduces additional errors during classification.

## 5 Conclusions

In this paper we presented two methods allowing the fusion of knowledge contained in one neural network into another without access to the original training data and without iterative learning. One method (Weights Summation) uses the statistical properties of weights in trained neural networks and is based on a simple procedure of weights summation or averaging. The other method (Elastic Weight Consolidation) tries to minimize the change of weights essential to the networks that are being fused. The methods were tested with feedforward shallow and deep convolutional ANNs on several classification tasks and show similar performance. Numerical experiments show that the methods for information fusion works even in a case when ANNs contain completely independent knowledge.

In this paper, we demonstrated the examples of knowledge fusion between a pair of ANNs, however, the developed methods can be easily extended for the fusion of several networks (Akhlaghi & Sukhov, 2018).

The fusion methods allow transfer and addition of new knowledge into pretrained networks that will help in reducing time and computational costs in deployment of neural networks, for example, in mobile platforms. In the future, these methods can help in developing a unified protocol of information exchange between neural networks. In addition to machine learning applications, further development of the methods of information transfer from one neural net to another will help in understanding the processes of communication between different brain areas.

The proposed methods for fused networks can be used at an initialization stage for further training. For example, the proposed method can be used for a parallel training of several networks in the case of large scale data streams (Zhu, Ikeda, Pang, Ban, & Sarrafzadeh, 2018) with their further fusion. The methods can be also used as a benchmark for the future knowledge exchange methods and fusion algorithms.